%% file: main.tex
\documentclass[10pt,twocolumn,letterpaper]{article}
\usepackage[pagenumbers]{cvpr} 

\definecolor{cvprblue}{rgb}{0.21,0.49,0.74}
\usepackage[pagebackref,breaklinks,colorlinks,allcolors=cvprblue]{hyperref}
\usepackage{float}
\usepackage{multirow}

\def\paperID{2925} 
\def\confName{CVPR}
\def\confYear{2025}
\title{MOS: Modeling Object-Scene Associations in Generalized Category Discovery}
\author{Zhengyuan Peng$^{1}$, 
	  ~Jinpeng Ma$^{3}$,
        ~Zhimin Sun$^{1}$,
        ~Ran Yi$^{1}$,
        ~Haichuan Song$^{2}$,
        ~Xin Tan$^{2}$\footnotemark[2]~,
        ~Lizhuang Ma$^{1,2}$\\
	$^1$Shanghai Jiao Tong University,
	$^2$East China Normal University,
    $^3$Chongqing  University\\
}

\begin{document}

\twocolumn[{%
\maketitle
\begin{figure}[H]
\hsize=\textwidth 
\centering
    \includegraphics[scale=0.52]{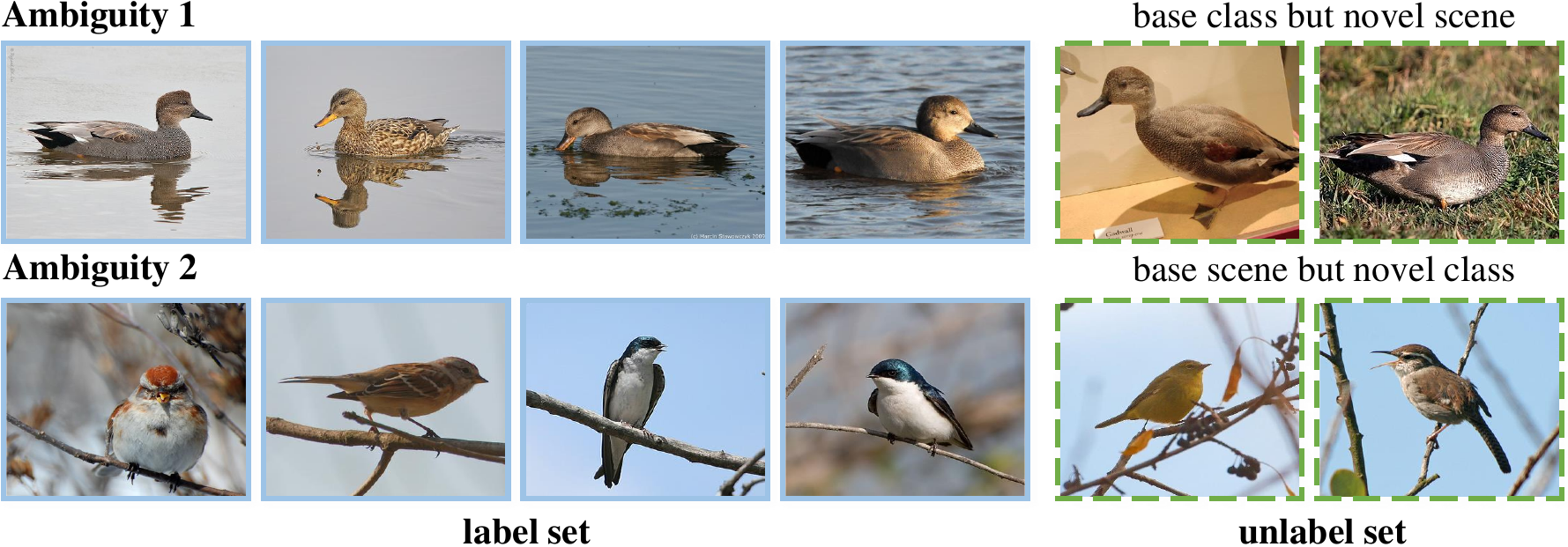}
    \caption{\small \textbf{Illustration of the ambiguity challenge.} The top row illustrates the potential for base objects in novel scenes to be perceived as novel categories. The bottom row shows misclassification risks when novel objects are placed in base scenes. Ambiguity challenge is the primary factor that leads to the misinterpretation of scene information.}
    \label{ambiguity}
    \vspace{1em}
\end{figure}
}]

{\renewcommand{\thefootnote}{\fnsymbol{footnote}}
\footnotetext[2]{Project Leader.}}

\begin{abstract}
Generalized Category Discovery (GCD) is a classification task that aims to classify both base and novel classes in unlabeled images, using knowledge from a labeled dataset. In GCD, previous research overlooks scene information or treats it as noise, reducing its impact during model training. However, in this paper, we argue that scene information should be viewed as a strong prior for inferring novel classes. We attribute the misinterpretation of scene information to a key factor: the \textbf{Ambiguity Challenge} inherent in GCD. Specifically, novel objects in base scenes might be wrongly classified into base categories, while base objects in novel scenes might be mistakenly recognized as novel categories. Once the ambiguity challenge is addressed, scene information can reach its full potential, significantly enhancing the performance of GCD models. To more effectively leverage scene information, we propose the \textbf{Modeling Object-Scene Associations (MOS)} framework, which utilizes a simple MLP-based scene-awareness module to enhance GCD performance. It achieves an exceptional average accuracy improvement of \textbf{4\%} on the challenging fine-grained datasets compared to state-of-the-art methods, emphasizing its superior performance in fine-grained GCD. The code is publicly available at \href{https://github.com/JethroPeng/MOS}{https://github.com/JethroPeng/MOS}.
\end{abstract}

\begin{figure*}[]
    \centering
    \includegraphics[scale=0.46]{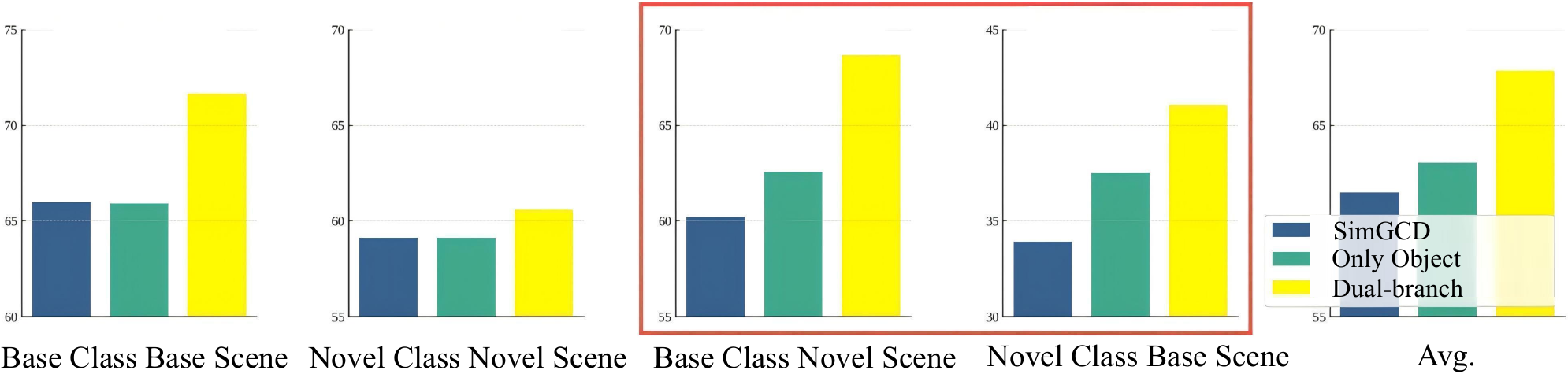}
    \caption{\small \textbf{Comparison of the performance on the 4 subsets of the CUB dataset.} The red box highlights scenarios involving two types of ambiguities. We compare the performance variations between the scene-removed image (green) and the original image (dark blue) as inputs. Our analysis reveals that the most significant performance improvement occurs in situations involving a novel-base relationship conflict between object and scene. In other situations, there is a slight decline in performance. Furthermore, employing dual-branch network for training (yellow) leads to a notable performance enhancement across four subsets.}
    \label{insights}
\end{figure*} 

\section{Introduction}\label{sec1}
Deep learning models in image recognition often face challenges when novel classes are introduced, due to their reliance on large annotated datasets and closed-world assumption. This limitation sparks interest in Generalized Category Discovery (GCD)~\cite{GCD}, which aims to classify novel categories from unlabeled data using knowledge from labeled datasets. GCD enables models to generalize to novel categories and adapt to diverse environments, overcoming the constraints of the closed-world assumption.

In GCD, existing approaches~\cite{DCCL,GCD,zhang2022promptcal,wang2024sptnet,Parametric_GCD,Learning_Semi-supervised_GCD, MetaGCD, IGCD, CLIP-GCD,cao2021open,han2021autonovel,fini2021unified,wang2023discover,zhao2023learning,chiaroni2023parametric,fei2022xcon,vaze2024no,rastegar2024learn} often overlook scene information (e.g., forest, ocean, sky) or treat it as noise that interferes with the generalization of classification models, seeking to minimize its influence through various techniques. This viewpoint has long been widely accepted, supported by a key experiment: performance notably improves in GCD when the scene is removed. However, we argue that the scene is not noise. Instead, in environments with limited information about novel classes, scene information can actually serve as a useful signal for inferring the class, providing strong prior knowledge that supports more accurate predictions.\par

This paper finds that scene information is often misinterpreted as noise due to \textbf{Ambiguity Challenge} inherent in GCD. Specifically, when a novel scene contains base objects, these objects are more likely to be misclassified as novel due to the difficulty in distinguishing between scene and object. Conversely, novel objects within a familiar scene may be erroneously classified into base categories. We conduct an observation experiment on the CUB dataset. In our experiments, we categorize scenes as novel or base class based on their presence in the labeled set. Based on the base and novel relationships between objects and scenes, we divide the dataset into four subsets. We find that the improvement from scene removal primarily occurs in situations where the base-novel relationship between the object and the scene conflict, as is shown in Fig.~\ref{insights}. We illustrate ambiguity challenge in Fig.~\ref{ambiguity}, which confirms our views.\par

Further experiments demonstrate that once the ambiguity challenge is addressed, the scene information can enhance object classification performance. Based on the scene-removed image and original image, we further employ a dual-branch network for joint learning, where the model can learn scene information through the contrastive differences. Compared to the performance of object image, results suggest that enabling the network to learn scene information achieves significant improvements in accuracy, outperforming the single-branch model across all four subsets. Results are shown in Fig.~\ref{insights}. In Sec.~\ref{sia}, we investigate whether the network effectively captures the scene information.\par

To more effectively leverage scene information, we propose the \textbf{Modeling Object-Scene Associations (MOS)} framework, which utilizes a simple MLP-based scene-awareness module to enhance GCD performance. In our framework, we employ a universal saliency segmentation model to perform zero-shot segmentation of objects and scenes. Their features are then extracted in the same feature space using a shared backbone for consistent representation. Scene-awareness module processes these two features for perceptual differentiation, extracting more effective criteria for categorization. Following extensive experimentation, \textbf{MOS} surpasses state-of-the-art in fine-grained GCD tasks, with only a minimal increase in network.  Notably, it achieves an impressive average accuracy of \textbf{65\%} on three parts of the Semantic Shift Benchmark (SSB), underlining its superior performance. Compared to the baseline, we achieve a \textbf{9\%} performance improvement.\par

Our research indicates that scene information provides more benefits than drawbacks in fine-grained GCD task. Reducing scene information introduces a performance bottleneck in scenarios with limited information. Effective mining of scene information can provide more comprehensive criteria for discovering different novel classes.\par

We summarize our contributions as follows:
\begin{enumerate}
    \item We challenge the traditional view that scene information interferes with GCD. Our experiments show that, when the ambiguity challenge is addressed, scene information can serve as a valuable prior for inferring novel categories,  improving fine-grained GCD performance.\par
    \item We propose a novel framework, \textbf{Modeling Object-Scene Associations (MOS)}, which incorporates a simple MLP-based scene-awareness module to effectively leverage scene information in GCD. Our approach significantly outperforms existing state-of-the-art methods on fine-grained datasets.\par
    \item We annotate the scene information in the CUB dataset to enable further exploration of scene impacts and support related research.
\end{enumerate}

\section{Related work}

\textbf{Semi-Supervised Learning}
leverages both labeled and unlabeled data to enhance model training, proving particularly valuable in situations where only a small proportion of the dataset is labeled. Basic methods in semi-supervised learning include self-training~\cite{lee2013pseudo, xie2020self}, consistency regularization~\cite{laine2016temporal, tarvainen2017mean}, and other approaches~\cite{berthelot2019mixmatch, sohn2020fixmatch, chen2023softmatch,chen2024beyond,chen2023boosting,feng2022dmt}.  Self-training consists of two stages: initial training on labeled data, followed by further training with pseudo-labels generated from unlabeled data. Consistency regularization ensures that predictions for augmented versions of the same data remain consistent. Another important variant of the semi-supervised learning is open-set semi-supervised learning~\cite{chen2020semi,Wang_2023_CVPR,li2023iomatch}, which focuses on discovering outliers in unlabeled datasets that do not correspond to any category in the labeled set. However, it generally does not distinguish between different types of outliers.\par
\noindent
\textbf{Generalized Category Discovery}
aims to classify both base and novel classes in unlabeled images, using knowledge from a labeled dataset. A seminal study~\cite{GCD} fine-tunes DINO features and categorizes different classes using semi-supervised k-means. SimGCD~\cite{Parametric_GCD} improves upon this by employing a parametric model instead of clustering algorithms, enhancing performance and robustness. DCCL~\cite{DCCL} introduces an innovative iterative framework that simultaneously estimates underlying visual concepts and learns their representations. PromptCAL~\cite{zhang2022promptcal} proposes a two-stage Contrastive Affinity Learning method using visual prompts. SPTNet~\cite{wang2024sptnet} employs a two-stage approach that iteratively fine-tunes both models and input data. In addition to these approaches, other methods~\cite{Learning_Semi-supervised_GCD, MetaGCD, IGCD, CLIP-GCD,cao2021open,han2021autonovel,fini2021unified,wang2023discover,zhao2023learning,chiaroni2023parametric,fei2022xcon,vaze2024no,rastegar2024learn,peng2023generalized,zheng2024textual,rastegar2024selex,choi2024contrastive,ma2024active,pu2024federated} also improve the performance of GCD from different perspectives. Most of them overlook scene information or treat it as noise. However, the underlying mechanisms and role of scene information in GCD remain underexplored.\par
\noindent
\textbf{Scene Information in classification.} Scene is an important topic in computer vision~\cite{ye2024spurious,WANG2023103646,li2018deep,Ma2022VisualRT,li2021ctnet,gong2023scene,sun2024uni,xie2024pig,sun2024image,tan2024mimir,tan2025edtalk,tan2024flowvqtalker}. The influence of scene information on classification performance remains controversial. Some studies~\cite{ye2024spurious} suggest that scene information primarily introduces spurious correlations, which can lead to noise and hinder the generalization of classification models, rather than contributing to meaningful classification signals. To this end, some models~\cite{caron2021emerging,oquab2023dinov2} are designed to prioritize object-centric information, utilizing attention modules that focus predominantly on objects. Concurrently, there is an emerging recognition of the significance of contextual scene information in enhancing classification performance, supported by recent research~\cite{WANG2023103646}. It shows that scene elements in an image provide crucial prior knowledge that aids in classification. The above studies are typically based on traditional classification tasks. In GCD task, the relationships become more complex with novel class. This paper attempts to delve into the impact of scene information in GCD.\par

\section{Methodology}
\subsection{Preliminary and Overview}
In Generalized Category Discovery (GCD), the dataset \( D \) comprises both a labeled subset \( D_l \) and an unlabeled subset \( D_u \), defined as \( D = D_l \cup D_u = \{(X, Y)\} \). The primary goal is to use the labeled dataset \( D_l = \{(X_l, Y_l)\} \) within the known label space \( C_l \), to effectively categorize samples in the unlabeled dataset \( D_u = \{(X_u)\} \). The label space of these unlabeled samples, \( C_u \), often includes \( C_l \) as a subset, \( C_l \subset C_u \). This setup presents a complex classification challenge as \( C_u \) includes additional categories unseen in \( C_l \).

In this section, we provide a detailed illustration of our method, \textbf{Modeling Object-Scene Associations (MOS)}. MOS takes two inputs: the original image \( X \) and the segmented object \( O \). In Sec.~\ref{32}, we first define the method for object extraction. Sec.~\ref{33} offers an overview of the MOS framework and describes its algorithmic flow. Sec.~\ref{34} delves into the scene-awareness module, a key component of the \textbf{MOS} framework, which enhances performance.

\subsection{Object-Scene Decoupling}
\label{32}
We utilize the universal zero-shot saliency segmentation model IS-Net~\cite{IS-Net} to separate objects from their scene. The segmentation process generates a saliency mask \( M 
\in \{0, 1\}^{m \times n} \), where foreground objects are represented by 1 and the background by 0. We extract object image \( O \) from the original image \( X \) using the formula:
\begin{equation}
O = X \cdot M + \mu \cdot (1-M), 
\end{equation}

\noindent
where \( \mu \) represents a filling value used to replace the scene pixels, \( m \) and \( n \) denote the width and height of the original image  \( X \), respectively.  In practice, we set  \( \mu \) to the mean pixel value of the image. This method is essential for minimizing the domain gap caused by missing scene parts.

\begin{figure*}[!t]
    \centering
    \vspace{1em}
\includegraphics[scale=0.65]{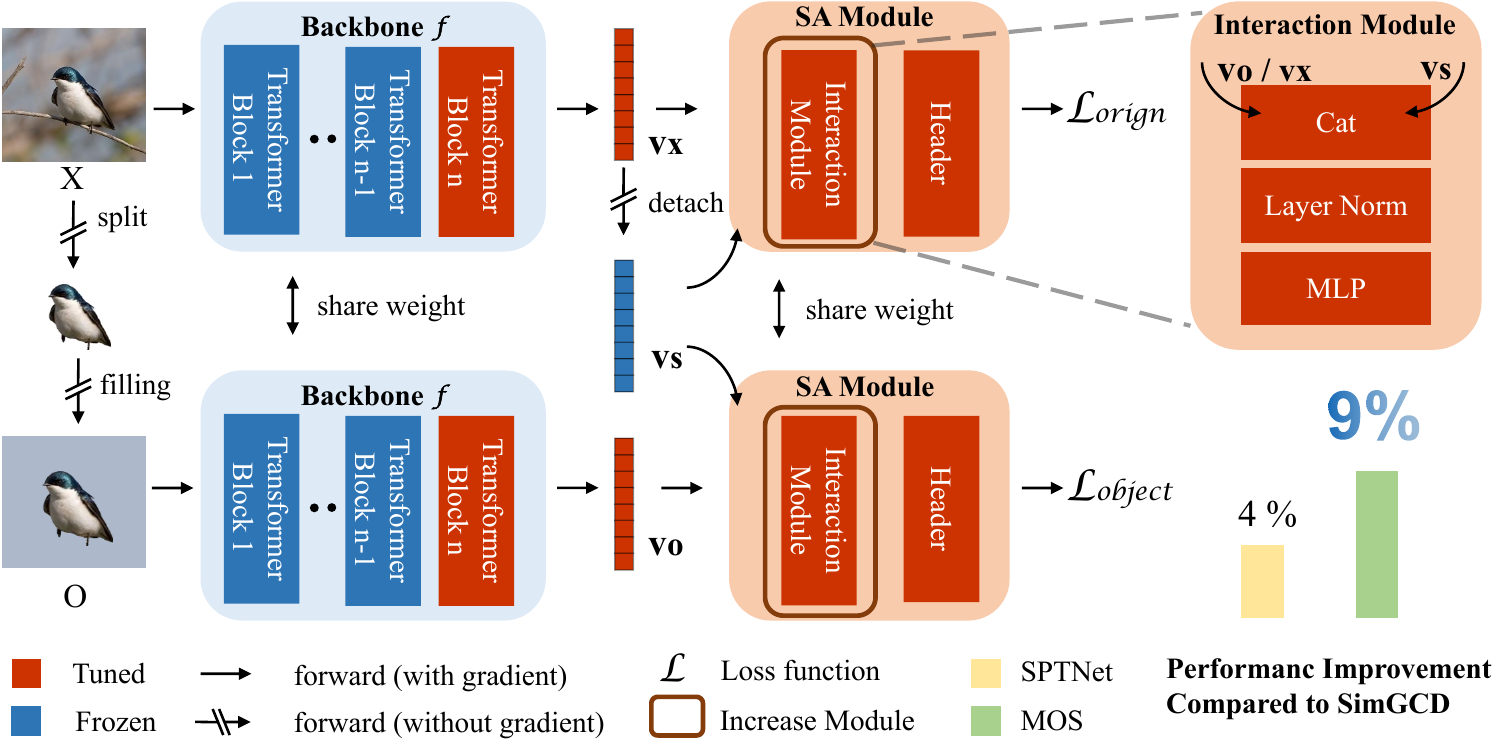}
    \caption{\small  \textbf{Modeling Object-Scene Associations (\textbf{MOS}) Framework.} The framework adopts a dual-branch design: one branch processes the original image, and the other handles the segmented object image. Both branches share two core components: the \textbf{Backbone} \( f \) and the \textbf{Scene-awareness Module} \( \theta \). During training, the original image is segmented using a universal saliency segmentation model to extract the object. The scene regions of the object image are then filled with the mean pixel value. Both the original and object images are fed into the Backbone. After extracting features \( v_o \) from the object image and \( v_x \) from the original image, the scene features \( v_s \) are obtained from \( v_x \). In the scene-awareness module, \( v_x \) and \( v_s \), as well as \( v_o \) and \( v_s \), interact to produce the classification output and compute the losses. During evaluation, only the output from the object branch is used.}
\end{figure*}

\subsection{Modeling Object-Scene Associations}
\label{33}

Scene information provides essential contextual knowledge for classification. Thus, we introduce the novel \textbf{MOS} framework, which effectively models object-scene associations with only a minimal increase in network. \textbf{MOS} framework includes two primary components: the backbone network \( f \) and the scene-awareness module \( \theta \). Compared to SimGCD~\cite{Parametric_GCD}, \textbf{MOS} introduces only minimal additional training parameters, specifically by adding a single training MLP in scene-awareness module.

The input to MOS consists of the original image and the image after object extraction. The method for object extraction is described in Sec.~\ref{32}. The shared backbone network extracts both object and scene features simultaneously within the same feature space. We use the information obtained from the object image as object features. As for the scene features, scene information \( S \), typically located at the image edges, is difficult for the DINO network to extract as standalone features. Given that the conditional entropy of the scene features is consistent with the conditional entropy of the original image feature under the same condition of object features ( The original image only consist of both scene and object components ) , we therefore replace the scene features with the original image features.\par

Our framework is divided into two branches: one processes the object parts \( O \) , and the other focuses on the original images \( X \). The overall pipeline is listed as follows:
\begin{equation}
\left.\begin{aligned}
v_o &= f(\text{O}) \\
v_x &= f(X) \\
v_s &:= v_x \\
\end{aligned}\right\}
\quad \text{Feature}
\quad \left.\begin{aligned}
\hat{y}_{\text{origin}} &= \theta(v_x, v_s) \\
\hat{y}_{\text{object}} &= \theta(v_o, v_s) \\
\end{aligned}\right\}
\quad \text{Output},
\end{equation}

\noindent
where $\hat{y}_{\text{origin}}$ and $\hat{y}_{\text{object}}$ dictate the output of two branch. In the original image branch, the primary distinction from the object image branch includes using the image features \( v_x \), instead of \( v_o \). \( v_x \) and \( v_s \) are interacted within the scene-awareness module to produce the classification output.\par

\begin{figure*}[]
    \centering
    \vspace{1em}
    \includegraphics[scale=0.58]{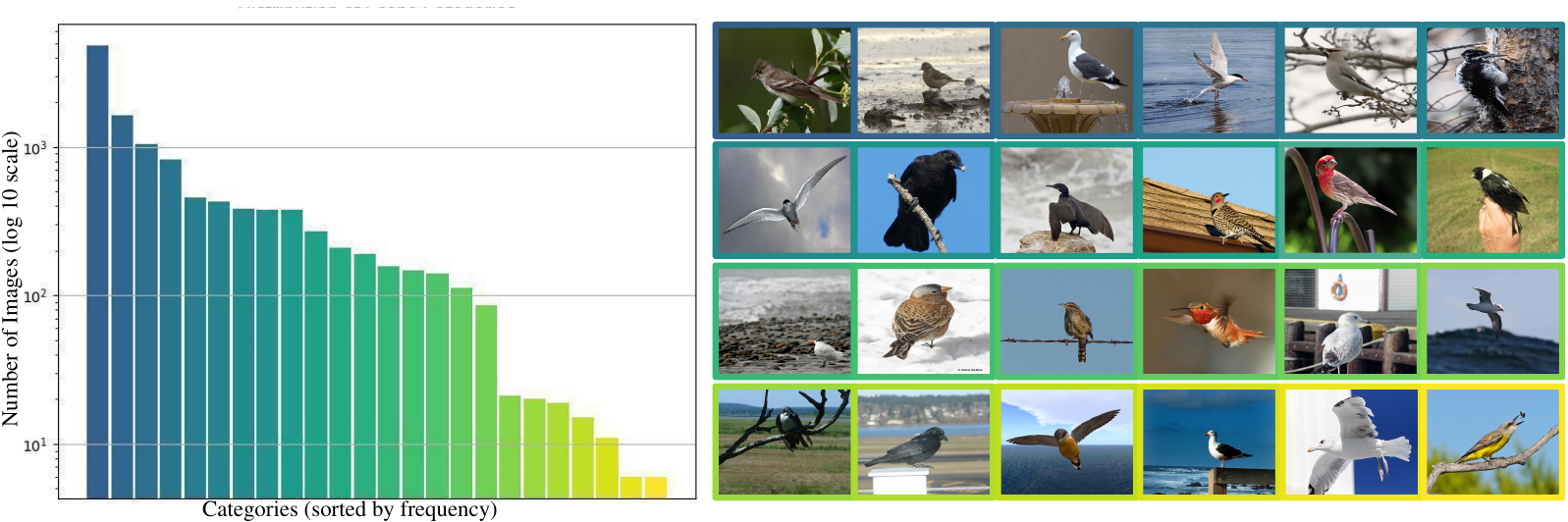}
    \caption{\small \textbf{Display of CUB Scene Information.} The left side shows category counts with a log-transformed y-axis to better illustrate the distribution. The right side shows category examples, illustrating the dataset's diversity.}
    \label{datasetshow}
\end{figure*} 

Two branches share the same backbone \( f \) and the SA module \( \theta \), but losses are calculated separately. For each branch, we compute losses for both labeled and unlabeled images. For labeled images, we calculate the supervised contrastive learning loss \( L_{\text{sup}}^{\text{nce}} \) and the classification loss \( L_{\text{sup}}^{\text{cls}} \). For unlabeled images, we assess the unsupervised contrastive learning loss \( L_{\text{un}}^{\text{nce}} \) and teacher-student cross-entropy loss \( L_{\text{un}}^{\text{cls}} \). \(L_{\text{sup}}^{\text{nce}} \) and \( L_{\text{un}}^{\text{nce}} \) pull same-class features closer and push different-class features in the feature space. \( L_{\text{sup}}^{\text{cls}} \) computes the cross-entropy between the output and labels, while \( L_{\text{un}}^{\text{cls}} \) enforces consistency regularization for unlabeled set. The combined loss \cite{Parametric_GCD} is given as follows:
\begin{equation}
 L_{\text{origin}}/L_{\text{object}} = (1 - \lambda) (L_{\text{un}}^{\text{nce}} + L_{\text{un}}^{\text{cls}}) + \lambda (L_{\text{sup}}^{\text{nce}} + L_{\text{sup}}^{\text{cls}}),
\end{equation}
where \( \lambda \) is a weighting factor that balances supervised and unsupervised losses. The overall loss \( L \) is given as follows:
\begin{equation}
L = \lambda_1 L_{\text{origin}} + \lambda_2 L_{\text{object}}, 
\end{equation}
where \( \lambda_1 \) and \( \lambda_2 \) are weighting factors that balance the influence of each branch. In practical applications, setting both \( \lambda_1 \) and \( \lambda_2 \) to 1 is a common strategy to ensure balanced training and help improve performance. During evaluation, both object and original images are input and processed by the backbone \( f \) to extract \( v_o \), \( v_x \), and \( v_s \). \( v_o \) and \( v_s \) pass through object branch's SA Module for output, while original image branch's SA Module is discarded.

\subsection{Scene-awareness Module}
\label{34}

In \textbf{MOS}, the features from the original image \(X\) are denoted as \(v_x\), and the features corresponding to the object \(O\) are represented as \(v_o\). The scene features \(v_s\) are also derived from the original image features \(v_x\). However, it introduces two challenges during training: 1) Replacing \(v_s\) with \(v_x\) leads to rapid variations in the feature space, making it difficult to capture meaningful information differences; 2) Over-optimization of \(v_x\) can cause an imbalance in the learning process, potentially compromising performance. To address these issues, we introduce a teacher network that remains fixed during training (i.e., The teacher output \( v_s \) is typically treated as a student output \( v_x \)  with no gradient propagation (detach) ). The teacher network outputs the scene features \(v_s\), thereby maintaining the stability of the features and alleviating the challenges associated with training.

We input these features into the scene-awareness module, a concise and efficient interaction design. It only includes an Interaction Module \( IM \) and Header. \( IM \) normalizes the merged features to reduce discrepancies between inputs from different branches. We only need to employ a Multilayer Perceptron (MLP) for interaction. The formula for the Interaction Module \( IM \) is as follows:

\begin{equation}
\small
\mathbf{IM}_i = \text{MLP}\!\left(\frac{\mathbf{v}_i \oplus \mathbf{v}_s}{\|\mathbf{v}_i \oplus \mathbf{v}_s\|}\right), \quad i \in \{o, x\},
\end{equation}

\noindent
where the feature vectors for the original image, object, and scene are denoted as $\mathbf{v}_x, \mathbf{v}_o$ and $\mathbf{v}_s$, respectively. These vectors are concatenated and subsequently normalized by their combined norm. At the final stage of the scene-awareness module, we handle output through a DINO header. To ensure fairness in comparisons, we remove one MLP layer from the DINO header to compensate for the modules we added. The final classification is determined using the Hungarian matching algorithm. 

\begin{table}[htp] 
\small 
\centering 
\caption{\textbf{Dataset Statistics.}} 
\begin{tabular}{l c c c c} 
\hline 
 &  \multicolumn{2}{c}{Label} & \multicolumn{2}{c}{Unlabel}\\
 \cmidrule(lr){2-3} \cmidrule(lr){4-5}
 & Num & Class  & Num & Class  \\ 
\hline 
CUB~\cite{welinder2010caltech} & 1.5K & 100 & 4.5K & 200 \\ 
Stanford Cars~\cite{krause20133d} & 2.0K & 98 & 6.1K & 196 \\ 
FGVC-Aircraft~\cite{maji2013fine} & 1.7K & 50 & 5.0K & 50 \\ 
Oxford-IIIT Pet~\cite{vedaldi2012cats} &0.9K&19&2.7K&37\\
\hline 
\end{tabular} 
\end{table}

\begin{table*}[]
\vspace{1em}
\centering
\caption{\textbf{Evaluation on the fine-grained datasets.} Values in bold indicate the top results.}
\label{fgd}
\setlength{\tabcolsep}{4.5pt} 
\begin{tabular}{llllcllclllcll}  
\hline
\multirow{2}{*}{Method} & \multirow{2}{*}{Venue/Year} & \multicolumn{3}{c}{CUB} & \multicolumn{3}{c}{Stanford Cars} & \multicolumn{3}{c}{FGVC-Aircraft}  & \multicolumn{3}{c}{Avg.} \\ \cline{3-14} 
                  &      & All & \multicolumn{1}{c}{Base} & \multicolumn{1}{c}{Novel} & All & \multicolumn{1}{c}{Base} & \multicolumn{1}{c}{Novel} & All & \multicolumn{1}{c}{Base} & \multicolumn{1}{c}{Novel} & All & Base & Novel \\ \hline
                k-means++~\cite{arthur2007k} & SODA/2007 & 34.3 & 38.9 & 32.1 & 12.8 & 10.6 & 13.8 & 12.9 & 12.9 & 12.8 & 20.0 & 20.8 & 19.6 \\
                RankStats+~\cite{han2021autonovel} & TPAMI/2021 & 33.3 & 51.6 & 24.2 & 28.3 & 61.8 & 12.1 & 27.9 & 55.8 & 12.8 & 29.8 & 56.4 & 16.4 \\
                UNO+\cite{fini2021unified} & ICCV/2021 & 35.1 & 49.0 & 28.1 & 35.5 & 70.5 & 18.6 & 28.3 & 53.7 & 14.7 & 33.0 & 57.7 & 20.5 \\
                GCD~\cite{GCD} & CVPR/2022 & 51.3 & 56.6 & 48.7 & 39.0 & 57.6 & 29.9 & 45.0 & 41.1 & 46.9 & 45.1 & 51.8 & 41.8 \\
                XCon~\cite{fei2022xcon} & BMVC/2022 & 52.1 & 54.3 & 51.0 & 40.5 & 58.8 & 31.7 & 47.7 & 44.4 & 49.4  & 46.8 & 52.5 & 44.0 \\
                ORCA~\cite{cao2021open} & ICLR/2022 & 36.3 & 43.8 & 32.6 & 31.9 & 42.2 & 26.9 & 31.6 & 32.0 & 31.4 & 33.3 & 39.3 & 30.3 \\
                DCCL~\cite{DCCL} & CVPR/2022 & 63.5 & 60.8 & 64.9 & 43.1 & 55.7 & 36.2 & - & - & - & - & - & - \\
                GPC~\cite{zhao2023learning} & ICCV/2023 & 52.0 & 55.5 & 47.5 & 38.2 & 58.9 & 27.4 & 43.3 &  40.7 & 44.8 & 44.5 & 51.7 & 39.9 \\
                PIM~\cite{chiaroni2023parametric} & ICCV/2023 & 62.7 & 75.7 & 56.2 &43.1 & 66.9 & 31.6 & -  &-  &- & - & - & - \\
                SimGCD~\cite{Parametric_GCD} & ICCV/2023 & 60.3 & 65.6 & 57.7 & 53.8 & 71.9 & 45.0 & 54.2 & 59.1 & 51.8 & 56.1 & 65.5 & 51.5 \\
                PromptCAL~\cite{zhang2022promptcal} & CVPR/2023 & 62.9 & 64.4 & 62.1 & 50.2 & 70.1 & 40.6 & 52.2 & 52.2 & 52.3 & 55.1 & 62.2 & 51.7 \\
                InfoSieve~\cite{rastegar2024learn} & NIPS/2023 & 69.4 & \textbf{77.9} & 65.2 & 55.7 & 74.8 & 46.4  & 56.3 & 63.7 & 52.5 & 60.5 & 72.1 & 54.7 \\
                $\mu$GCD~\cite{vaze2024no} & NIPS/2023 & 65.7 & 68.0 & 64.6 & 56.5 & 68.1 & 50.9 & 53.8 & 55.4 & 53.0 & 58.7 & 63.8 & 56.2 \\
                GCA~\cite{otholt2024guided} & WACV/2024 & 68.8 &73.4 & 66.6 &54.4 & 72.1 & 45.8 &52.0 &57.1& 49.5&58.4&67.5&54.0 \\
                SPTNet~\cite{wang2024sptnet} & ICLR/2024 & 65.8 & 68.8 & 65.1 & 59.0 & 79.2 & 49.3 & 59.3 & 61.8 & 58.1 & 61.4 & 69.9 & 57.5 \\
                LeGCD~\cite{cao2024solving} & CVPR/2024 &63.8 & 71.9 & 59.8 & 57.3 & 75.7 & 48.4 & 55.0 & 61.5 & 51.7 &58.7 & 69.7& 53.3\\
                \hline
                MOS (our) & & \textbf{69.6} & 72.3 & \textbf{68.2} & \textbf{64.6} & \textbf{80.9} & \textbf{56.7} & \textbf{61.1} & \textbf{66.9} & \textbf{58.2} & \textbf{65.1} & \textbf{73.4} & \textbf{61.0} \\
                \hline
\end{tabular}
\end{table*}

\section{Experiments}
\subsection{Datasets} 
We assess our approach using various fine-grained benchmarks tailored for image recognition. We employ the CUB~\cite{welinder2010caltech}, Stanford Cars~\cite{krause20133d}, and FGVC-Aircraft~\cite{maji2013fine} datasets, all part of the Semantic Shift Benchmark (SSB). For a more comprehensive evaluation, we also use Oxford-IIIT Pet~\cite{vedaldi2012cats}.  The dataset division adheres to established protocols. We subsample $|C_l|$ seen (labeled) classes and follow the previous work~\cite{GCD,rastegar2024learn}, selecting 50\% of samples the datasets for the labeled set \(D_l\), with the remaining samples constituting the unlabeled set \(D_u\).\par

For our observation experiment, we annotate the scene information of the CUB~\cite{welinder2010caltech} dataset. There are 24 different categories, which exhibit a clear long-tail distribution, as shown in Fig.~\ref{datasetshow}. We categorize unlabeled CUB data into four subsets: Base Class with Base Scene, Novel Class with Base Scene, Base Class with Novel Scene, and Novel Class with Novel Scene. The division between novel class and base classes follows the settings of the SSB benchmark. For the definition of scenes, categories that are absent or appear very infrequently in the label set are designated as novel scenes, while all others are considered base scenes. This dataset can also be used for other tasks related to scene studies.\par

\subsection{Evaluation Protocol}
We assess model performance using clustering accuracy (ACC), aligning with standard benchmarks. This involves matching ground truth labels, \( y_i \), with predicted labels, \( \hat{y}_i \), and calculating the ACC as follows:
\begin{equation}
\text{ACC} = \frac{1}{|D_u|} \sum_{i=1}^{|D_u|} \mathbf{1}(y_i = G(\hat{y}_i)),
\end{equation}
where \( G \) represents the optimal permutation that aligns the predicted labels with the ground truth most accurately.

\subsection{Implementation Details}
We propose the \textbf{MOS} framework, which builds upon the SimGCD baseline and incorporates a pre-trained ViT-B/16 backbone for robust feature extraction. Our data augmentation and parameter learning strategies follow those of prior studies~\cite{Parametric_GCD, wang2024sptnet}, ensuring consistency and comparability in performance evaluations. Specifically, our data augmentation pipeline includes scaling, horizontal flipping, cropping, and color jittering, which introduces a variety of training samples. For saliency segmentation, we use the IS-Net~\cite{IS-Net} model, a universal zero-shot saliency segmentation framework. We train with a batch size of 128 for 200 epochs, starting with an initial learning rate of 0.1, which decays according to a cosine schedule for each dataset. In addition, we set the balancing factor \( \lambda = 0.35 \) and the temperature parameters \( \tau_u = 0.07 \) and \( \tau_c = 1.0 \), in line with the settings in~\cite{Parametric_GCD}. The temperature values for classification losses are set to \( \tau_t = 0.07 \) and \( \tau_s = 0.1 \). We fine-tune the training hyperparameters on the FGVC-Aircraft. For FGVC-Aircraft, we apply a weight decay of 5e-4 and perform a warm-up of \( \tau_t \) from 0.04 to 0.07 after 20 epochs. All experiments run on a single NVIDIA GeForce RTX 4090 GPU.\par

\begin{table}[t]
 \centering
        \caption{\textbf{Evaluation on the Oxford-IIIT Pet datasets.} Values in bold indicate the top results.}
        \begin{tabular}{lcllc}
            \hline
            {Method} 
                                    & Venue/Year & All       & Base       & Novel     \\ \hline
            k-means++~\cite{arthur2007k}            & SODA/2007 & 77.1 & 70.1 & 80.7 \\ 
            GCD~\cite{GCD}                          & CVPR/2022 & 80.2 & 85.1 & 77.6 \\ 
            XCon~\cite{fei2022xcon}                 & BMVC/2022 & 86.7 & 91.5 & 84.1 \\ 
            DCCL~\cite{DCCL}                        & CVPR/2022 & 88.1 & 88.2 & 88.0 \\
            SimGCD~\cite{Parametric_GCD}            & ICCV/2023 & 88.8 & 84.9  & 90.9 \\ 
            InfoSieve~\cite{rastegar2024learn}      & NIPS/2023 & 91.8 & \textbf{92.6}  & 91.3\\     
            \hline
            MOS (our)                               &  & \textbf{93.2}&89.5&\textbf{95.1} \\ 
             \hline
        \end{tabular}
\end{table}

\subsection{Results}

The results from Tab.~\ref{fgd} demonstrate that our MOS method achieves a significant leap forward in fine-grained image classification. Specifically, our approach outperforms the baseline SimGCD with an average improvement of \textbf{9\%} on three parts of the Semantic Shift Benchmark (SSB). In comparison, our method also exhibits superior performance, achieving an average enhancement of \textbf{5\%}. The results indicate that scene information provides significant improvements in class inference for both base and novel classes.\par

We also assess \textbf{MOS} on Oxford-IIIT Pet dataset, ImageNet-100. The results clearly show that our approach significantly improves performance in comparison to previous methods. Specifically, our method outperforms the baseline SimGCD by \textbf{4.4\%} in terms of accuracy.

\subsection{Ablation Study}
To better understand the performance improvements achieved by our proposed \textbf{MOS} framework, we conduct an  ablation study on the CUB dataset, with detailed results shown in Tab.~\ref{ablation}. We decompose the MOS framework into its individual components and evaluate each one separately to quantify their contributions. The results demonstrate that each component contributes positively to the overall performance, benefiting both base and novel classes.\par

\begin{table}[t]
\centering
\caption{\textbf{Comparison of framework components.} \textbf{Object} refers to using object images for traditional training. \textbf{MOS} denotes the model excluding all components except the scene-awareness module. \textbf{SA Module} refers to the scene-awareness module itself. Each component contributes significantly to the overall performance improvement.}
\label{ablation}
\begin{tabular}{ccccccc}
\toprule
\textbf{Object} & \textbf{MOS} & \textbf{SA Module} & \textbf{All} & \textbf{Base} & \textbf{Novel} \\
\midrule
- & - & - & 61.50 & 65.70 & 59.40 \\
\checkmark & - & - & 63.06 & 64.71 & 62.23 \\
\checkmark & \checkmark & - & 67.86 & 70.78 & 66.40\\
\checkmark & \checkmark & \checkmark & \textbf{69.57} & \textbf{72.31} & \textbf{68.20} \\
\bottomrule
\end{tabular}
\end{table}

\begin{figure}[t]
    \centering
    \includegraphics[scale=0.3]{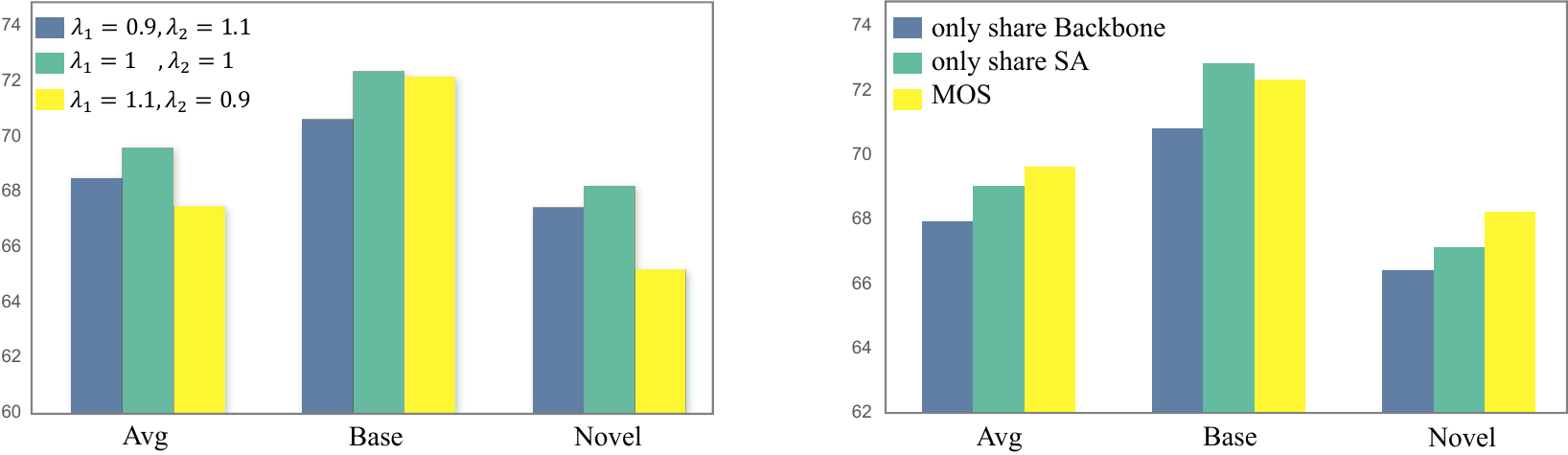}
    \caption{\small \textbf{Ablation Study on Parameters (left) and Shared Weights (right) on the CUB dataset.} The ablation study on parameters shows that the method is robust to parameter changes. The ablation study on shared weights reveals that shared weights primarily influence the inference of novel classes}
    \label{abf}
\end{figure}

We conduct a parameter ablation study, as shown in Fig.~\ref{abf}, to systematically assess the impact of different branch losses on model performance. The results demonstrate the effectiveness of the proposed balancing strategy and indicate that small deviations from the optimal configuration result in minimal performance degradation.\par 

We explore the use of shared weights across multiple stages, as illustrated in Fig.~\ref{abf}. Results show that employing shared weights enhances both model performance, highlighting the advantages of this approach in improving both effectiveness and resource utilization. The ablation study on shared weights reveals that shared weights have a significant impact on the performance with novel classes, indicating that effective extraction of scene features plays a crucial role in improving the inference of novel classes.

\begin{figure}[t]
    \centering
    \includegraphics[scale=0.48]{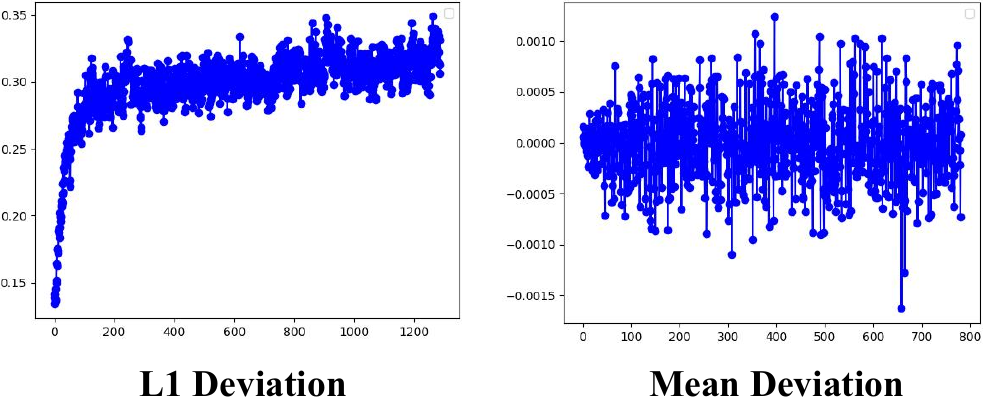}
    \caption{\textbf{Variation of mean deviation and L1 deviation between global image features and object features during training.} The x-axis represents the training iterations. The mean deviation (right) fluctuates within a small range (± 0.001), suggesting that the global image and object features form a unified and stable prototype. In contrast, the L1 deviation (left) increases notably, especially in the early stages of training, indicating the network's growing ability to distinguish between the global image and object features as it learns to perceive scene differences.}
\end{figure} 

\begin{figure*}[t]
    \centering
    \vspace{1em}
    \includegraphics[scale=0.4]{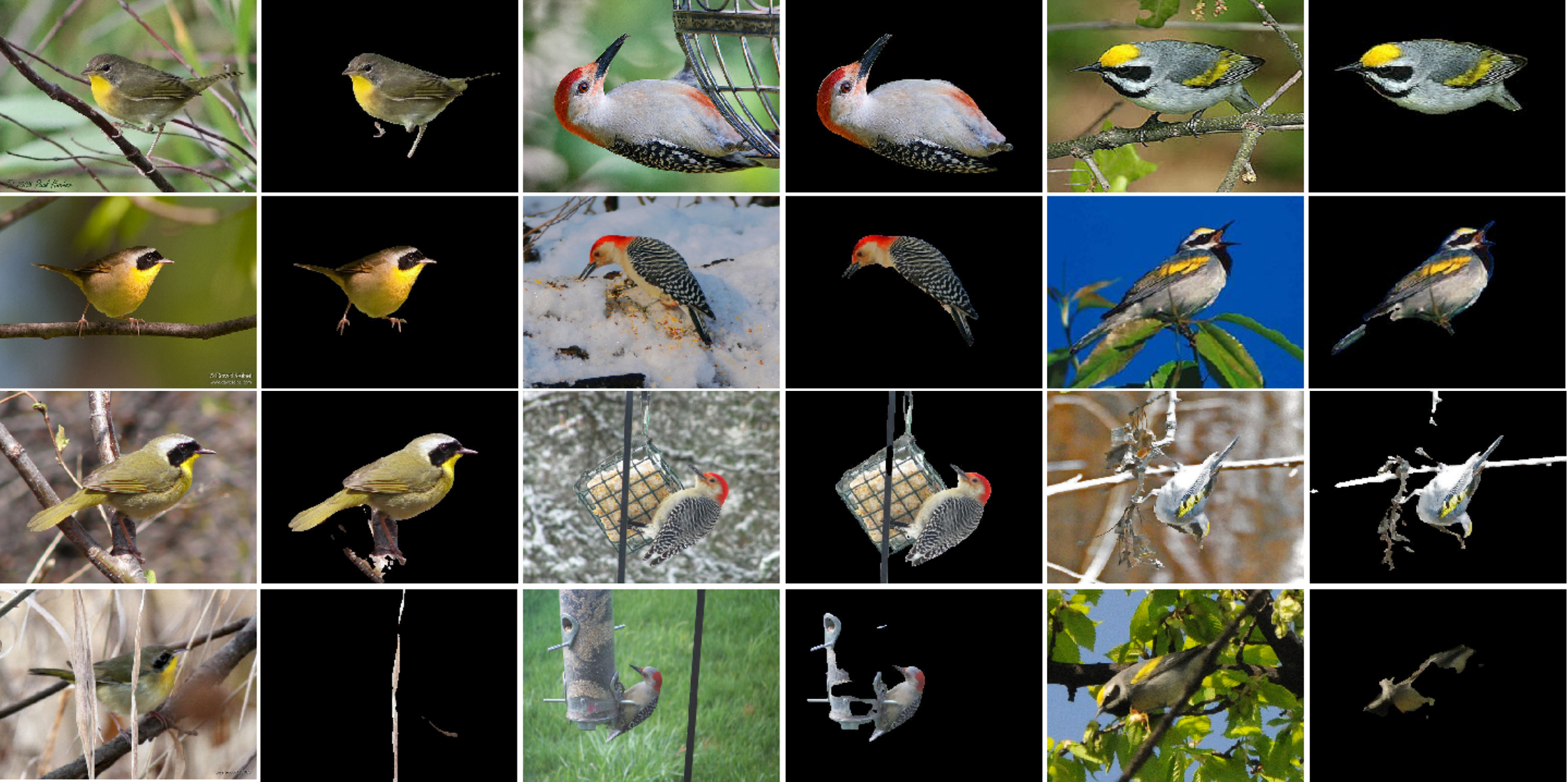}
    \caption{\small \textbf{Segmentation Visualization on the CUB dataset.} It displays the segmentation results of IS-Net. Although the model produces several poor cases (Line 4), this does not undermine the overall effectiveness of our approach, highlighting its robustness against occasional failures.}
    \label{v2}
\end{figure*} 

\subsection{Scene Information Analysis}
\label{sia}

We analyze whether the network has learned scene features by examining the differences between global image features and corresponding object features. Specifically, we track the mean deviation and L1 deviation of these differences throughout training. Mean deviation represents the central deviation between the global image features and the object features. If the mean deviation fluctuates within a narrow range, it indicates that the feature prototypes of the global image and object are similar, suggesting that the network has formed stable and consistent class prototypes. L1 deviation measures the relative deviation between the global image features and the object features. Compared to object-centric feature extractors like DINO, we observe that the L1 deviation increases as training progresses. This indicates that the network is learning to distinguish the differences between the global image features and the object features. Together, these analyses show that the network is effectively learning to capture scene features, as evidenced by the growing divergence between global and object-specific features during training.\par

\subsection{Visualization}
\begin{figure}[t]
    \centering
    \includegraphics[scale=0.33]{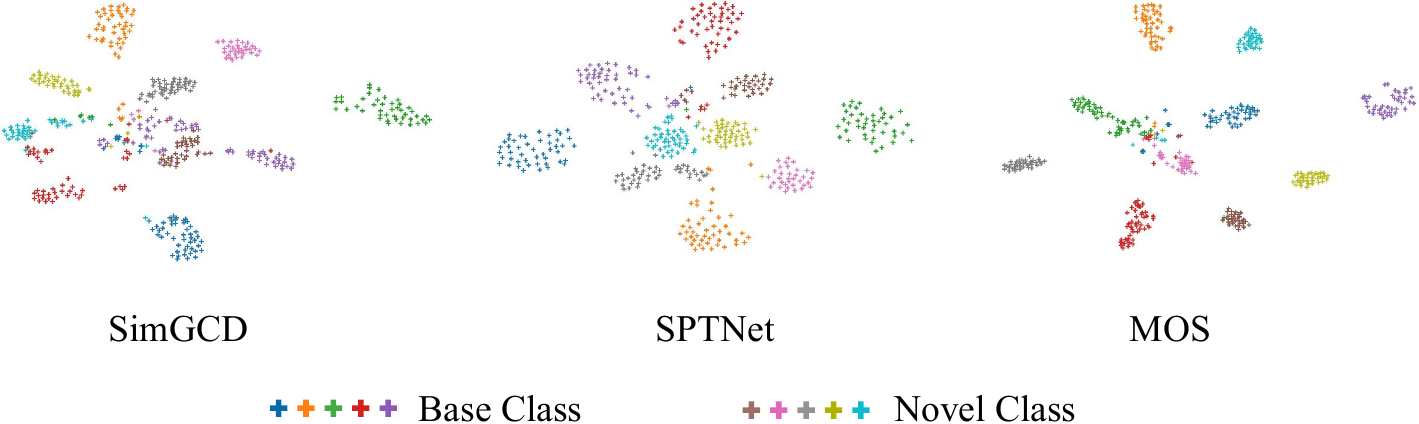}
    \caption{\small \textbf{The t-SNE visualization of representations of 10 classes randomly sampled from CUB dataset}. The results indicate that MOS is beneficial for maximizing inter-class margins.}
    \label{v1}
\end{figure}

\textbf{t-SNE.}
The t-SNE visualization, showcased in Fig.~\ref{v1}, highlights the strengths of our method, showcasing significantly tighter clustering and more distinct category separation than both SimGCD and SPTNet. It proves that the incorporation of scene information effectively increases the inter-class margin.\par

\noindent
\textbf{Segmentation.}
Our current saliency segmentation model, IS-Net, is exclusively used for zero-shot segmentation inference, with no additional training.  A common concern is whether our method is overly sensitive to segmentation performance fluctuations. We illustrate results of segmentation in Fig.~\ref{v2}. The saliency segmentation model produces many poor cases, yet our model still performs excellently, showcasing its robustness.\par

\section{Conclusion}
In this paper, we present a novel perspective on the role of scene information in Generalized Category Discovery (GCD), challenging the assumption that scenes act as noise in model training. Our findings demonstrate that scene information can serve as a crucial source of prior knowledge, enhancing the model’s ability to classify both base and novel categories. We identify the Ambiguity Challenge as a key factor contributing to the misinterpretation of scene information in GCD tasks, where the overlap of objects from base and novel categories in different scenes can lead to misclassification. After addressing this challenge, we show that scene information can significantly improve classification performance. To leverage this insight, we propose the Modeling Object-Scene Associations (MOS) framework, which incorporates a simple MLP-based scene-awareness module that effectively distinguishes between scene and object features. Our experiments on fine-grained datasets demonstrate that MOS outperforms existing state-of-the-art methods. This result underscores the importance of integrating scene information into the GCD framework, rather than dismissing it as irrelevant.

\paragraph{Discussion.}
Our method achieves excellent performance in fine-grained datasets. However, it struggles to extract effective object information in extremely low-resolution settings, such as CIFAR 10/100.

\section{Acknowledgment}
This work is supported by the National Natural Science Foundation of China (62302167, U23A20343, 62472282, 72192821, 62302297), Shanghai Sailing Program (23YF1410500, 22YF1420300),  Chenguang Program of Shanghai Education Development Foundation Shanghai Municipal Education Commission (23CGA34), and Young Elite Scientists Sponsorship Program by CAST (2022QNRC001).

{
    \small
    \bibliographystyle{ieeenat_fullname}
    \bibliography{main}
}
\end{document}